\documentclass{article}

\usepackage[final]{nips_2018}

\usepackage[utf8]{inputenc} % allow utf-8 input
\usepackage[T1]{fontenc}    % use 8-bit T1 fonts
\usepackage{hyperref}       % hyperlinks
\usepackage{url}            % simple URL typesetting
\usepackage{booktabs}       % professional-quality tables
\usepackage{amsfonts}       % blackboard math symbols
\usepackage{nicefrac}       % compact symbols for 1/2, etc.
\usepackage{microtype}      % microtypography

\setcitestyle{numbers}
\setcitestyle{square}
\setcitestyle{comma}
\usepackage{graphicx}

\title{Conversational Intent Understanding for Passengers in Autonomous Vehicles}

\author{
  Eda Okur\\
  Intel Labs\\
  Anticipatory Computing Lab \\
  Hillsboro, OR 97124 \\
  \texttt{eda.okur@intel.com} \\
  \And
  Shachi H. Kumar \\
  Intel Labs\\
  Anticipatory Computing Lab \\
  Santa Clara, CA 95054 \\
  \texttt{shachi.h.kumar@intel.com} \\
  \And
  Saurav Sahay \\
  Intel Labs\\
  Anticipatory Computing Lab \\
  Santa Clara, CA 95054 \\
  \texttt{saurav.sahay@intel.com} \\
  \And
  Asli Arslan Esme \\
  Intel Labs\\
  Anticipatory Computing Lab \\
  Hillsboro, OR 97124 \\
  \texttt{asli.arslan.esme@intel.com} \\
  \And
  Lama Nachman \\
  Intel Labs\\
  Anticipatory Computing Lab \\
  Santa Clara, CA 95054 \\
  \texttt{lama.nachman@intel.com} \\
}

\begin{document}

\maketitle

\section{Introduction}
Understanding passenger intents and extracting relevant slots are important building blocks towards developing a contextual dialogue system responsible for handling certain vehicle-passenger interactions in autonomous vehicles (AV). When the passengers give instructions to AMIE (Automated-vehicle Multimodal In-cabin Experience), the agent should parse such commands properly and trigger the appropriate functionality of the AV system. In our AMIE scenarios, we describe usages and support various natural commands for interacting with the vehicle. We collected a multimodal in-cabin data-set with multi-turn dialogues between the passengers and AMIE using a Wizard-of-Oz scheme. We explored various recent Recurrent Neural Networks (RNN) based techniques and built our own hierarchical models to recognize passenger intents along with relevant slots associated with the action to be performed in AV scenarios. Our experimental results achieved F1-score of 0.91 on utterance-level intent recognition and 0.96 on slot extraction models. 

\section{Methodology}
\label{method}

Our AV in-cabin data-set includes 30 hours of multimodal data collected from 30 passengers (15 female, 15 male) in 20 rides/sessions. 10 types of passenger intents are identified and annotated as: \textit{Set/Change Destination}, \textit{Set/Change Route} (including turn-by-turn instructions), \textit{Go Faster}, \textit{Go Slower}, \textit{Stop}, \textit{Park}, \textit{Pull Over}, \textit{Drop Off}, \textit{Open Door}, and \textit{Other} (turn music/radio on/off, open/close window/trunk, change AC/temp, show map, etc.). Relevant slots are identified and annotated as: \textit{Location}, \textit{Position/Direction}, \textit{Object}, \textit{Time-Guidance}, \textit{Person}, \textit{Gesture/Gaze} (this, that, over there, etc.), and \textit{None}. In addition to utterance-level intent types and their slots, word-level intent keywords are annotated as \textit{Intent} as well. We obtained 1260 unique utterances having commands to AMIE from our in-cabin data-set. We expanded this data-set via Amazon Mechanical Turk and ended up with 3347 utterances having intents. The annotations for intents and slots are obtained on the transcribed utterances by majority voting of 3 annotators.

For slot filling and intent keywords extraction tasks, we experimented with seq2seq LSTMs and GRUs, and also Bidirectional LSTM/GRUs. The passenger utterance is fed into a Bi-LSTM network via an embedding layer as a sequence of words, which are transformed into word vectors. We also experimented with GloVe, word2vec, and fastText as pre-trained word embeddings. To prevent overfitting, a dropout layer is used for regularization. Best performing results are obtained with Bi-LSTMs and GloVe embeddings (6B tokens, 400K vocab size, dim 100).

For utterance-level intent detection, we experimented with mainly 5 models: (1) Hybrid: RNN + Rule-based, (2) Separate: Seq2one Bi-LSTM + Attention, (3) Joint: Seq2seq Bi-LSTM for slots/intent keywords \& utterance-level intents, (4) Hierarchical + Separate, (5) Hierarchical + Joint. For (1), we extract intent keywords/slots (Bi-LSTM) and map them into utterance-level intent types (rule-based via term frequencies for each intent). For (2), we feed the whole utterance as input sequence and intent-type as single target. For (3), we experiment with the joint learning models \cite{hakkani-2016, asr-liu-2016, zhang-2016} where we jointly train word-level intent keywords/slots and utterance-level intents (adding <BOU>/<EOU> terms to the start/end of utterances with intent types). For (4) and (5), we experiment with the hierarchical models \cite{h-zhou-2016, h-meng-2017, wen-2018} where we extract intent keywords/slots first, and then only feed the predicted keywords/slots as a sequence into (2) and (3), respectively.

\section{Experimental Results}
\label{exp_res}

The slot extraction and intent keywords extraction results are given in Table~\ref{T1} and Table~\ref{T2}, respectively. Table~\ref{T3} summarizes the results of various approaches we investigated for utterance-level intent understanding. Table~\ref{T4} shows the intent-wise detection results for our AMIE scenarios with the best performing utterance-level intent recognizer.

\begin{table}
  \caption{Slot Extraction Results (10-fold CV)}
  \label{T1}
  \centering
  \begin{tabular}{*2c}
    \toprule
    \textbf{Slot Type} & \textbf{F1} \\
    \toprule
    Location & 0.95 \\
    Position & 0.95 \\
    Person & 0.97 \\
    Object & 0.89 \\
    Time Guidance & 0.85 \\
    Gesture & 0.92 \\
    None & 0.98 \\
    \midrule
    \textit{AVERAGE} & \textit{\textbf{0.96}} \\
    \bottomrule
  \end{tabular}
\end{table}

\begin{table}
  \caption{Intent Keyword Extraction Results (10-fold CV)}
  \label{T2}
  \centering
  \begin{tabular}{*2c}
    \toprule
    \textbf{Keyword Type} & \textbf{F1} \\
    \toprule
    Intent & 0.94 \\
    Non-Intent & 0.99 \\
    \midrule
    \textit{AVERAGE} & \textit{\textbf{0.98}} \\
    \bottomrule
  \end{tabular}
\end{table}

\begin{table}[h]
  \caption{Utterance-level Intent Recognition Results (10-fold CV)}
  \label{T3}
  \centering
  \begin{tabular}{lc}
    \toprule
    %\multicolumn{1}{c}{Model Type}                   \\
    %\cmidrule(r){1-2}
    \textbf{Utterance-level Intent Detection Models} & \textbf{F1} \\
    %\midrule
    \toprule
    Hybrid-1: RNN + Rule-based (\textit{intent keywords}) & 0.86 \\
    Hybrid-2: RNN + Rule-based (\textit{intent keywords \& slots}) & 0.90\\
    \midrule
    Separate-1: Seq2one Bi-LSTM & 0.88 \\
    Separate-2: Seq2one Bi-LSTM + Attention (withContext) & 0.89\\
    \midrule
    Joint: Seq2seq Bi-LSTM (\textit{intent keywords \& slots \& utterance-level intent types}) & 0.88 \\
    \midrule
    Hierarchical \& Separate-1 & 0.90 \\
    Hierarchical \& Separate-2 (Separate-1 + Attention) & 0.90 \\
    \midrule
    Hierarchical \& Joint & \textbf{0.91} \\
    \bottomrule
  \end{tabular}
\end{table}

\begin{table}[!t]
  \caption{Intent-wise Performance Results of Utterance-level Intent Recognition Models: Hierarchical \& Joint (10-fold CV)}
  \label{T4}
  \centering
  %\begin{tabular}{{c}l{c}l{c}l{c}l{c}}
  \begin{tabular}{*3c}
    \toprule
    %\multicolumn{1}{c}{Model Type}                   \\
    \textbf{AMIE Scenario} & \textbf{Intent Type} & \textbf{F1} \\
    %\cmidrule(r){3-4} \\
    %\midrule
    \toprule
     & Stop & 0.92 \\
    Finishing the Trip & Park & 0.94 \\
    Use-cases & PullOver & 0.96 \\
     & DropOff & 0.95 \\
    \midrule
    Set/Change & Set/ChangeDest & 0.89 \\
    Destination/Route & Set/ChangeRoute & 0.88 \\
    \midrule
    Set/Change & GoFaster & 0.88 \\
    Driving Behavior/Speed & GoSlower & 0.88 \\
    \midrule
    Others & OpenDoor & 0.95 \\
    (Door, Music, A/C, etc.) & Other & 0.82 \\
    \midrule
     & \textit{AVERAGE} & \textit{\textbf{0.91}} \\
    \bottomrule
  \end{tabular}
\end{table}

\section{Conclusion}
After exploring various recent Recurrent Neural Networks (RNN) based techniques, we built our own hierarchical joint models to recognize passenger intents along with relevant slots associated with the action to be performed in AV scenarios. Our experimental results outperformed certain competitive baselines and achieved overall F1-scores of 0.91 for utterance-level intent recognition and 0.96 for slot extraction tasks.

%\bibliography{amie_wiml_nips_2018}

\end{document}